 \documentclass[sn-basic]{sn-jnl}

\usepackage{graphicx}%
\usepackage{multirow}%
\usepackage{amsmath,amssymb,amsfonts}%
\usepackage{amsthm}%
\usepackage[title]{appendix}%
\usepackage{xcolor}%
\usepackage{textcomp}%
\usepackage{manyfoot}%
\usepackage{algorithm}%
\usepackage{algorithmicx}%
\usepackage{algpseudocode}%
\usepackage{listings}%
\usepackage{natbib}
\usepackage{adjustbox}
\usepackage{caption}
\usepackage{float}
\usepackage{tabularx}
\usepackage{amssymb}
\usepackage{amsmath}
\usepackage{graphicx}
\usepackage{tabularx}
\usepackage{multirow}
\usepackage{float}
\usepackage{caption}
\usepackage{adjustbox}
\usepackage{subfig}
\usepackage{afterpage}
\usepackage{xcolor}
\usepackage{hyperref}
\usepackage[T1]{fontenc}

\theoremstyle{thmstyleone}%
\theoremstyle{thmstyletwo}%

\theoremstyle{thmstylethree}%

\begin{document}

\title[LiDAR Loop Closure Detection using Semantic Graphs with Graph Attention Networks]{LiDAR Loop Closure Detection using Semantic Graphs with Graph Attention Networks}


\author*[1]{Liudi Yang}
\email{liudiyang1998@gmail.com}
 \author[1]{Ruben Mascaro}
 \author[3]{Ignacio Alzugaray}
 \author[2]{Sai Manoj Prakhya}
 \author[1]{Marco Karrer}
 \author[2]{Ziyuan Liu}
 \author[1]{Margarita Chli}

 \affil[1]{{Vision for Robotics Lab, ETH Zurich and University of Cyprus},  {Switzerland and Cyprus}}
             
 \affil[2]{{Huawei Munich Research Center}, {Germany}}

 \affil[3]{{Department of Computing, Imperial College London}, {UK}}


\abstract{In this paper, we propose a novel loop closure detection algorithm that uses graph attention neural networks to encode semantic graphs to perform place recognition and then use semantic registration to estimate the 6 DoF relative pose constraint. Our place recognition algorithm has two key modules, namely, a semantic graph encoder module and a graph comparison module. The semantic graph encoder employs graph attention networks to efficiently encode spatial, semantic and geometric information from the semantic graph of the input point cloud. We then use self-attention mechanism in both node-embedding and graph-embedding steps to create distinctive graph vectors. The graph vectors of the current scan and a keyframe scan are then compared in the graph comparison module to identify a possible loop closure. Specifically, employing the difference of the two graph vectors showed a significant improvement in performance, as shown in ablation studies. Lastly, we implemented a semantic registration algorithm that takes in loop closure candidate scans and estimates the relative 6 DoF pose constraint for the LiDAR SLAM system. Extensive evaluation on \textcolor{red}{public datasets} shows that our model is more accurate and robust, achieving 13\% improvement in maximum F1 score on the SemanticKITTI dataset, when compared to the baseline semantic graph algorithm. For the benefit of the community, we open-source the complete implementation of our proposed algorithm and custom implementation of semantic registration at \textbf{\textit{\href{https://github.com/crepuscularlight/SemanticLoopClosure}{https://github.com/crepuscularlight/SemanticLoopClosure}}}.}





\maketitle

\section{Introduction}\label{sec1}

Simultaneous Localization and Mapping (SLAM) plays a crucial role in enabling autonomous mobile robots to explore and navigate unknown environments. One of its fundamental challenges is the accumulation of drift caused by state estimation errors in the front-end odometry, which can lead to globally inconsistent maps. To address this issue, loop closure detection algorithms are developed that identify revisited places and help in reducing the accumulated drift by adding a 6 DoF pose constraint in pose graph or nonlinear factor-graph-based LiDAR SLAM systems. \textcolor{red}{There has been continual innovation in loop closure detection algorithms by employing the learning capability in latest AI advancements \cite{PointNetVLAD, LCDNet, SGPR,LPDNet,TransLiDAR} to improve their accuracy and real-time performance.}

Classical (learning-free) LiDAR-based loop closure detection algorithms use heuristic and handcrafted methods to reduce a large raw point cloud to a distinctive and descriptive multi-dimensional vector. Moreover, most of these handcrafted methods also have a carefully designed metric specific to the descriptor to compare their similarity. \textcolor{red}{ Many such handcrafted global feature descriptors such as M2DP~\cite{M2DP}, ScanContext~\cite{ScanContext} and its variants~\cite{ISC,scancontext++} are highly sensitive to design parameters, type of the LiDAR used, i.e., 32, 64 or 128 channel, the pose of the LiDAR (horizontal or inclined) and lastly perceptual disturbances such as occlusion and rotation. }

 \textcolor{red}{In recent years, deep learning-based loop closure detection algorithms have gained significant traction due to their ability to be custom-trained for the target environment, taking into account the specific installation pose and type of LiDAR used. Early representative methods such as PointNetVlad~\cite{PointNetVLAD} and LCDNet~\cite{LCDNet} already exhibited promising accuracy by learning generalizable and discriminative point cloud descriptors. However, they process the raw point clouds as a whole and need substantial computing power in deployment, thereby compromising real-time performance. }

A more recent work, SGPR~\cite{SGPR} proposed a less computationally intensive approach that explicitly incorporates a semantic graph as the underlying representation in an attempt to better mimic how the real world scene actually looks semantically. To this end, SGPR takes the instance segmentation result of the point cloud as the input, creates semantic graphs, and encodes spatial and semantic information into lightweight graph embeddings. These graph embeddings are matched in a graph-graph interaction module, which is a graph-matching neural network that treats loop detection as a graph comparison problem. SGPR \cite{SGPR}, being one of the first graph-based approaches for place recognition, did not encode comprehensive information from the semantic graph (missed out the geometric information), used a much simpler EdgeConv module \cite{SGPR} to create node embeddings and there is still some scope to improve their graph-graph interaction module as proposed in our paper. 

Recently graph attention networks (GAT) \cite{GAT} and graph similarity computation methods \cite{simgnn} have shown significant improvements on how graphs can be encoded, compared and can be made learnable in an end-to-end fashion. Specifically, graph attention networks use learnable linear transformation instead of simple scalar values to aggregate neighboring features, offering better graph encoding than EdgeConv, as used in SGPR \cite{SGPR}. This essentially allows the multi-head attention in GATs to learn from multiple subspaces, thus encoding complex relationships between graph nodes, offering a significant boost in performance. Next, the seminal paper on Transformers \cite{attention} introduced self-attention mechanism \cite{attention}, where the relationship between different elements of the input sequence can be learned effectively to reason about the underlying complex relations in training data. 

Inspired by SGPR \cite{SGPR}, GAT \cite{GAT}, self-attention \cite{attention} and SimGNN \cite{simgnn}, we have developed an enhanced graph-based loop closure detection that overcomes many drawbacks and uses the latest techniques to effectively encode a semantic graph, resulting in a significant boost in the performance. We propose a two-stage approach consisting of a semantic graph encoder and a graph comparison module. 
\begin{itemize}
    \item As our {\textit{{first contribution}}}, we enhance SGPR by designing a semantic graph encoder that uses graph attention networks to encode spatial, semantic and geometric information of semantic graph as opposed to SGPR's limited information and simpler encoding. 
    \item Our {\textit{{second contribution}}} is to use self-attention mechanism in node embedding and graph embedding steps to encode complex underlying relationships, essentially creating more distinctive graph vectors. 
    \item As our {\textit{{third contribution}}}, we show that employing the difference of the input graph vectors in the graph comparison module to perform classification offers a significant boost in the performance, as opposed to direct usage of graph vectors, as in SGPR \cite{SGPR}. 
    \item Our {\textit{{final contribution}}} is to open source our work at \textbf{\textit{\href{https://github.com/crepuscularlight/SemanticLoopClosure}{\small{https://github.com/crepuscularlight/SemanticLoopClosure}}}} that consists of semantic graph encoder module, graph comparison module, custom implementation of semantic registration for 6 DoF pose estimation to foster further research in this direction. 
\end{itemize}

Exhaustive experiments and ablation studies on public datasets prove the increased accuracy and robustness of both our semantic place recognition network and semantic registration algorithm compared to other methods from the state of the art. In addition, we demonstrate that both modules can run in real time with minimal memory and compute requirements, making them an ideal choice to integrate into existing SLAM frameworks.

\section{Related Work}\label{sec2}

We review previous works on traditional and learning-based 3D place recognition algorithms, and related graph neural networks that can function as backbones to extract representative features from graphs, \textcolor{red}{underscoring the limitations of existing approaches and setting the stage for the improvements introduced by our method.}

\subsection{3D Place Recognition}

\textbf{Traditional methods} reduce a raw point cloud with millions of points into a multi-dimensional vector using meticulously designed methods. Mostly, these extracted descriptors can be compared using Euclidean distance or specific handcrafted metrics to find a close match, essentially representing a place match/revisit. Magnusson et. al. \cite{NDT} has developed NDT, a histogram-based feature descriptor, exploiting normal distribution representation to describe 3D surfaces and an evaluation metric for scene matching. In 2013, Bosse et al. \cite{KeypointVoting} presented a keypoint voting mechanism to achieve fast matching between the current scan and database scans while estimating the matching thresholds/hyper-parameters by fitting a parametric model to the underlying distribution. M2DP \cite{M2DP} first projects 3D point clouds into a 2D plane to generate density signatures and uses corresponding concatenated singular vectors as descriptors.

SegMatch \cite{segmatch} is one of the first approaches that extracted descriptors from the clustered segments of the raw point clouds and used a geometric verification step to find a correct match. The approaches, where high-level geometric clustering and semantic/feature description are used for matching, generally achieve high accuracy and are more robust in loop closure detection. Scan Context \cite{ScanContext} initiated the trend to directly use point clouds without calculating histograms to create a global descriptor. It uses an encoding function that stores condensed information in spatial bins along radial and azimuthal directions to generate more distinctive global descriptors. While vanilla Scan Context \cite{ScanContext} literally only stores the maximum height in each bin, its variants encode more effective and representative information in the bins including detected intensity \cite{ISC}, semantic labels of point clouds \cite{SSC} and subcontexts \cite{scancontext++} boosting the performance of the descriptor. 

\textbf{Learning-based methods} essentially have the advantage of being able to custom train them for the target environment with specific robot/sensor setup, thus enabling them to offer better performance in particularly complex environments. PointNetVLAD \cite{PointNetVLAD} leveraged deep neural networks to retrieve large-scale scenes by using PointNet \cite{PointNet} as the backbone and NetVLAD \cite{NetVLAD} to aggregate learned local features, while outputting global descriptors for matching. SegMap \cite{SegMap} has proposed to learn data-driven leveraging on 3D point cloud variance of each cluster/segment, but the innate 3D CNN architecture comes with considerable computational burden. 

To enrich local geometric details, LPD-Net \cite{LPDNet} resorts to an adaptive backbone to aggregate local information into the global descriptors. The core of the local information extraction module is to fuse the nearest neighbors' information from feature space and Cartesian space. MinkLoc3D \cite{Mink} proposed a simple neural network to process sparse voxelized point clouds based on sparse 3D CNN. By quantizing the raw point clouds and employing sparse convolution, it achieves a similar inference speed to other multilayer-perceptron-based algorithms while maintaining high precision. LCDNet \cite{LCDNet} adopted end-to-end architecture simultaneously accomplishing place recognition and 6-DOF pose estimation. The shared 3D voxel CNN is used to extract features for the two-head output of place recognition and pose estimation. SGPR \cite{SGPR} on the other hand, converted the place recognition into a graph-matching problem by deeming every instance as a node and designed an efficient graph neural network to infer the similarity while exhibiting excellent robustness on mainstream datasets. 

\subsection{Graph Neural Networks for Feature Extraction}

Here, we review a few related works from graph neural networks that encode graphs into a distinctive feature vector, as it is the core idea of SGPR and our proposed algorithm. Inspired by the tremendous success of CNN in the computer vision field, plenty of methods transplant convolution into graph structures. ConvGNN \cite{ConvGNN} developed a graph convolution based on the spectral graph theory. GCN \cite{GCN} proposed a semi-supervised way to implement an efficient variation of graph convolution that utilizes first-order approximation in the spectral field. In order to boost the flexibility of graph neural networks, DGCNN \cite{DGCNN} adopted EdgeConv to dynamically aggregate features from nearest neighbors which are more suitable for high-level features in the graph. 
The appearance of ResGCN \cite{ResGCN} showed the superior advantage by adding residual networks and introducing large-scale architectures.

Another branch of GNN originates from the self-attention mechanism \cite{attention}. The graph attention networks \cite{GAT} (GAT) proposed a novel idea of leveraging self-attention layers to calculate learnable weights from neighbors to encode complex relations with more parameters from multiple subspaces of graph nodes. This work opened up a completely new direction of working with graph data and has been applied to various fields. Brody et al. \cite{Attentive} explored the innate mechanism of GAT and modified the order of operations to overcome the original limitation of static attention. 

In this paper, we take SGPR as a baseline and further enhance its architecture with graph attention networks, extract additional geometric information of instances, use self-attention mechanism to encode complex relations and develop a more 
discriminative graph comparison module that offers a significant performance boost with a lower model size and faster inference.

\section{Method}\label{sec3}
\begin{figure*}[h!]
  \begin{center}
  \includegraphics[width=\linewidth]{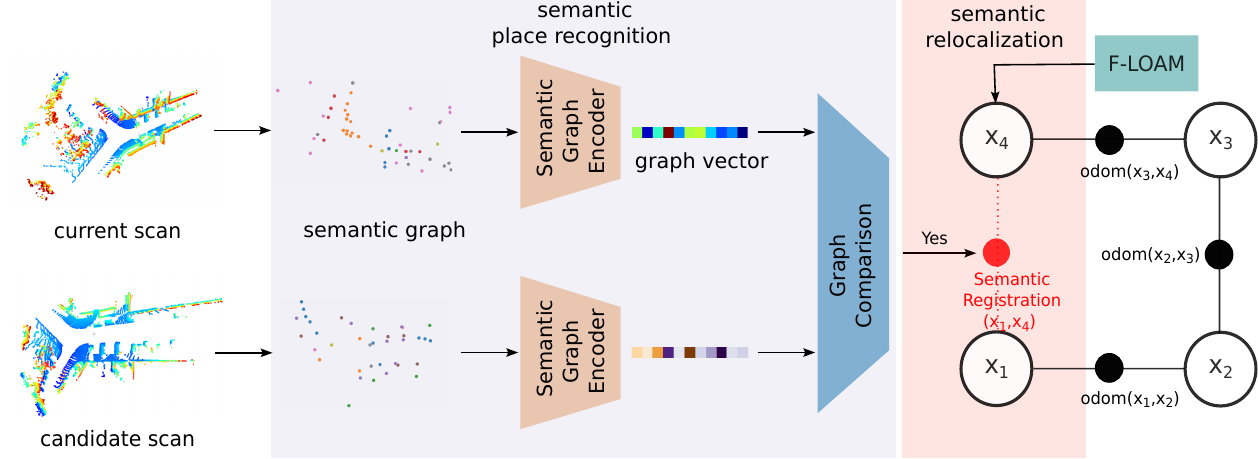} 
  \end{center}

  \caption{The high-level workflow of the proposed semantic graph based loop closure system integrated into a SLAM framework. The proposed loop closure algorithm takes two semantically segmented point clouds as input, which are converted to semantic graphs. After that, semantic graph encoders are deployed to compress them into graph vectors. Finally, the graph comparison module predicts the similarity of the two loop candidates. When the similarity exceeds a specific threshold, a pose constraint is estimated using semantic registration, which is added to the pose graph for trajectory optimization.}
  \label{fig:PRPipeline}
\end{figure*}

\subsection{System Overview}
Our proposed system's pipeline is shown in Fig. \ref{fig:PRPipeline}. F-LOAM \cite{FLOAM} is used as the front-end LiDAR odometry to provide the pose estimation for every incoming LiDAR scan. This pose is regarded as a node in the pose graph for nonlinear optimization of the whole trajectory. The relative pose according to odometry is added as a factor between the current node and its previous node. Specifically, when the semantic graph-based place recognition module finds potential loop candidates successfully, another constraint calculated by semantic registration is inserted into the pose graph between the corresponding nodes. This pose graph with both odometry constraints and loop closure constraints is optimized to get the final consistent and accurate 3D map of the environment. 
There are two components in the proposed loop closure back-end, first, a semantic place recognition module that generates loop closure candidates and second, a semantic relocalization to calculate the relative 6 DoF pose. 

\subsection{Semantic Place Recognition}
\label{PR}

The semantic place recognition module identifies if the robot has come back to a previously traversed location, and this in quantitative terms, provides a set of possible loop closure candidates. In order to judge whether two LiDAR scans are collected from the same place, we design this module with three parts:
\begin{enumerate}
    \item Construction of semantic graphs from input point clouds
    \item Semantic graph encoder module to extract distinctive graph vectors from semantic graphs
    \item Graph comparison module to estimate the similarity of two graph vectors

\end{enumerate}
If the predicted similarity is higher than a certain threshold, then the scan pair is regarded as a potential loop closure candidate for semantic registration and for further pose graph optimization.   

\subsubsection{Semantic Graph}

We adopt a similar strategy to SGPR \cite{SGPR} to construct semantic graphs from raw semantic point clouds, however, we add additional geometric information by encoding detected bounding boxes to enhance the performance. The nodes of the constructed semantic graph encode semantic labels of detected objects, their local centroid coordinates and bounding boxes. One-hot encoding is used as the embedding function for semantic information, which eliminates the ordinal influence of semantic labels. The encoded semantic labels and centroids $(x_k,y_k,z_k)$ of instances represent the spatial distribution of semantic objects and the topological relationship of those instances in the current scene. By adding the proposed bounding boxes, we enhance the semantic graph with additional geometric information representing the object's size and boundaries. \textcolor{red}{We have explored the following three possible ways of encoding the geometric information about the instances}:
\begin{itemize}
    \item FPFH \cite{fpfh} - Classical 3D feature descriptor
    \item PointNet \cite{PointNet} - Deep learning based 3D feature extractor 
    \item Bounding box (top left, bottom right) points of the instance
\end{itemize}
\textcolor{red}{Through quantitative evaluation(Table \ref{tab:geoCompare} in Section \ref{sec:geo_ablation}), we finalized on adding bounding boxes as an additional source of geometric information to the constructed semantic graph nodes, boosting the performance, with no extra computational overhead while bounding boxes were readily available through instance segmentation. }
\begin{figure*}[h!]
  \begin{center}
  \includegraphics[width=1\linewidth]{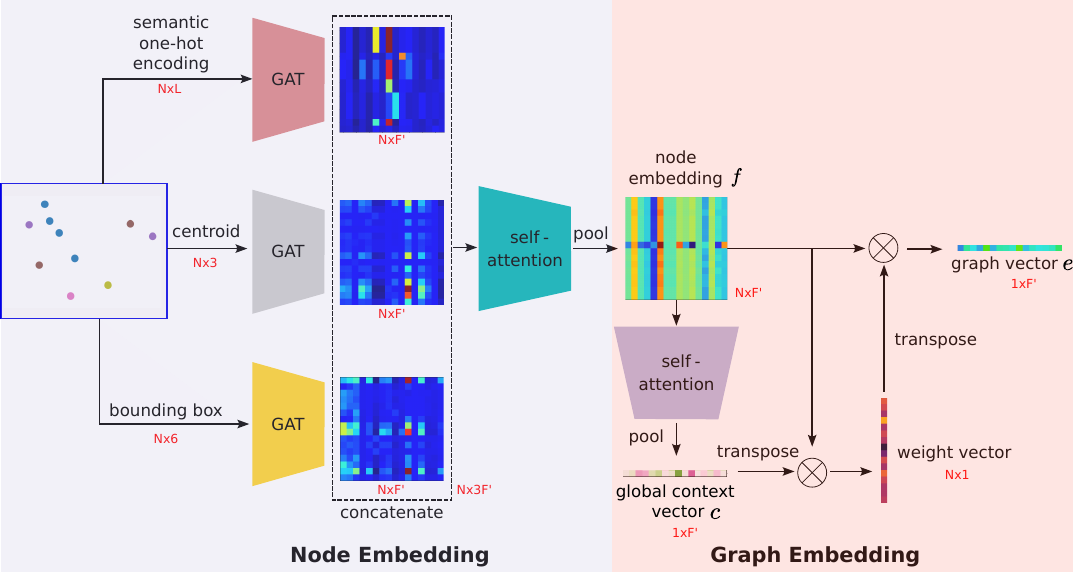} 
  \end{center}

  \caption{The architecture of the proposed semantic graph encoder. The semantic graphs created from input point clouds are passed through three GATs to extract contextual spatial, semantic and geometric features. These features are then concatenated and passed through a self-attention module to produce a node embedding $f$. Another self-attention module operated on the node embedding $f$ to learn a global context vector $c$. We finally project the node embedding $f$ into the global context vector $c$ to obtain corresponding node weights and use them to calculate the final graph vector $e$. }
  \label{fig:NodeEmbedding}
\end{figure*}
\subsubsection{Semantic Graph Encoder Module}
The semantic graph encoder is designed to convert the semantic graphs into representative graph vectors. In this process, these graph vectors become more distinguishable than the input semantic graphs, while representing them in less number of dimensions. The semantic graph encoder has two steps: extracting feature matrices from graphs (node embedding) and compressing feature matrices into vectors (graph embedding). 

\textbf{Node Embedding: } Given a graph with multiple nodes containing unique information, we encode the information of considered node and also the contextual information from its neighbouring nodes. In SGPR model, this is realized by the EdgeConv from DGCNN \cite{DGCNN}, which skips the time-consuming step of building adjacent matrices for every input graph and instead searches for \textit{k} nearest neighbor (kNN) nodes in the convolution operation. This made DGCNN more flexible and robust while reducing the computational overhead. Instead of using EdgeConv as in SGPR to encode neighbourhood node information, we observe that graph attention networks (GATs) can extract more comprehensive features that encode complex underlying relationships between neighborhood nodes by learning from multiple subspaces through multi-head attention. Hence, we propose to replace the EdgeConv backbone with GATs and subsequently modify the next steps in the whole pipeline.

The detailed illustration of semantic graph encoder is shown in Fig. \ref{fig:NodeEmbedding}. The semantic graph encoder has three branches emanating from the input semantic graph, that correspond to semantic labels, centroids and bounding boxes. For a considered branch, the input semantic graph can be denoted as $\mathbf{h}=\left\{\vec{h}_1, \vec{h}_2, \ldots, \vec{h}_N\right\}, \vec{h}_i \in \mathbb{R}^F$ where $N$ represents the number of nodes and $F$ is the feature dimension, for example, $F = 3$ for centroid and $F = 6$ for bounding box as shown in Fig. \ref{fig:NodeEmbedding}. 

To aggregate neighbourhood node information, we first find the difference between the current node and neighbouring node, $\vec{h}_i - \vec{h}_j$ and then concatenate this difference with $\vec{h}_i$, resulting in $\vec{h}_i \|  (\vec{h}_i-\vec{h}_j)$, essentially doubling its dimensionality. We perform this concatenation of $\vec{h_i}$ with $\vec{h_i}-\vec{h_j}$ to combine contextual information between $\vec{h_i}$ and $\vec{h_j}$. We then use a learnable matrix $W \in \mathcal{R}^{F' \times 2F} $ to transform $\vec{h}_i \|  (\vec{h}_i-\vec{h}_j)$ and estimate the attention-based weights $\alpha_{ij}$ as shown below 
\begin{equation}
\alpha_{i j}=\frac{\exp \left(\operatorname{LeakyReLU}\left(\overrightarrow{\mathbf{a}}^T\mathbf{W}\left[ \vec{h}_i \|  (\vec{h}_i-\vec{h}_j)\right]\right)\right)}{\sum_{k \in \mathcal{N}_i} \exp \left(\operatorname{LeakyReLU}\left(\overrightarrow{\mathbf{a}}^T\mathbf{W} \left[ \vec{h}_i \|  (\vec{h}_i-\vec{h}_k)\right]\right)\right)}
\end{equation}

where $\overrightarrow{\textbf{a}}$ is the learnable attention vector to reduce dimensions, $\|$ denotes concatenation and $\mathcal{N}_i$ is the global id of the \textit{k} nearest neighbor nodes to node $i$. 
\textcolor{red}{In our experiments, we used $k=10$ for the nearest neighbors parameter, in order to ensure consistency across all evaluations.}

As compared to vanilla/classical GAT \cite{GAT} that uses all of the nodes, our proposed \textit{k}-NN search reduces the computational costs in terms of both training and inference drastically. We used the LeakyReLU activation function with slope 0.2 to enhance the learning accuracy of neural network and alleviate the dead neuron issues during training. The $i$-th row of features extracted using GAT can be represented as 
\begin{equation}
\vec{h}_i^{\prime} = \underset{z=1}{\overset{Z}{\|}} \sigma\left(\sum_{j \in \mathcal{N}_i} \alpha_{i j}^z \mathbf{W}_z \left[ \vec{h}_i \|  (\vec{h}_i-\vec{h}_j)\right]\right)
\end{equation}
where $\sigma$ is a nonlinear function, $\|$ represents concatenation, $\alpha^z_{ij}$ are normalized attention coefficients computed by the $z$-th
attention vector ($\overrightarrow{\textbf{a}}^z$), and $\textbf{W}_z \in \mathcal{R}^{\frac{F'}{Z}\times 2F}$ is the $z$-th learnable linear transformation. In this formula, we apply the multi-head attention with head number $Z$. In this way, different heads can concentrate on different subspaces of features $\vec{h}_i \|  (\vec{h}_i-\vec{h}_j)$, boosting the expressiveness of the enhanced GATs. 

After extracting features individually from the semantic, spatial and geometric branches, we employed a self-attention module to fuse them together (Fig. \ref{fig:NodeEmbedding}). Our intention is to not only interact with neighboring nodes in one branch but also aggregate information from different branches thereby yielding a more comprehensive node-embedding representation. Self-attention mechanism is innately suitable for determining different weights in a sequence. The corresponding output of node embedding $f \in \mathcal{R}^{N \times F'}$ is calculated as 
\begin{equation}
f=pooling(softmax(\frac{Q(x)K^T(x)}{\sqrt{d_k}}V(x)))\end{equation}
where $Q$, $K$, $V$ are respectively the query, key and value mapping functions, $x \in \mathcal{R}^{N \times {3F'}}$ is the concatenated feature from three branches and $d_k$ is the dimension of keys.
Going through a self-attention module,  
the original graph containing three separate branches of information from different nodes gets converted to a single node embedding matrix ($f$). The next step is to compress this node embedding ($f$) into a graph vector ($e$).

\textbf{Graph Embedding: } Compressing graphs into fixed length vectors that encode the information from all nodes, is essential to compare two graphs efficiently and enable many downstream applications. We propose to use a self-attention module to learn a global context vector $c \in \mathcal{R}^{F'}$ from node embedding $f$, instead of a much simpler approach in SimGNN, to efficiently capture useful information from $f$. The node embedding matrix $f$ is passed through the self-attention module producing a stack of auxiliary vectors, which can be represented as  
\begin{equation}attention(f)=(u_1,\dots,u_N)^T \in \mathcal{R}^{N \times F'}
\end{equation}
where $u_i \in \mathcal{R}^{F'}$ represents the auxiliary vector for $i$-th row in the $attention(f)$. The learnable global context vector $c$ is estimated by pooling the auxiliary vectors, similar to SimGNN \cite{simgnn} as shown below
\begin{equation}
c=tanh(\frac{1}{N} \sum^{N}_{i=1}u_i)
\end{equation}
We then obtain the weight vector as shown in Fig.\ref{fig:NodeEmbedding}, representing the similarity between the node embedding $f$ and global context vector $c$ by calculating their inner product. After converting the weight vector into $[0,1]$ via a sigmoid function, we finally estimate the graph vector $e$ as the inner product of node embedding $f$ and weight vector as shown below,

\begin{equation}
e=\sum^N_{i=1}sigmoid(f_{i}^Tc)f_i
\end{equation}
where $f_i$ represents the $i$-th row of node embedding $f$.
This graph vector $e$ essentially compresses and represents the input semantic graph by encoding all the nodes and their spatial, semantic and geometric information in to a $F'$-dimensional vector. In our experiments, we create a 32-dimensional graph vector, and we found that changing the graph vector's dimension from 16-64 dimensions did not bring any noticeable changes to the performance.

\subsubsection{Graph Comparison Module}
\begin{figure*}[t]
  \begin{center}
  \includegraphics[width=1\textwidth]{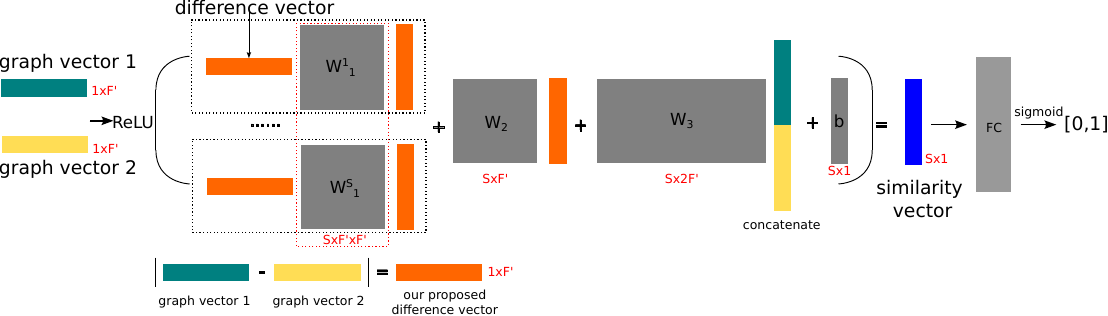} 
  \end{center}

  \caption{Overview of the graph comparison module. We propose a relative difference vector (shown in orange color) as the absolute value of the difference between two graph vectors. The similarity vector is leart based from first-order and second-order difference vectors, and concatenated graph vectors. This similarity vector is then passed through fully connected layers to predict the similarity value between two input graph vectors. }
  \label{fig:GraphInteraction}
\end{figure*}

The similarity between two graph vectors $e_1,e_2$ can be predicted by another neural network that can comprehend the resemblance and amplify the difference. We specifically enhance the performance of this module by adding an additional relative difference term, defined as $d=\lvert e_1-e_2 \rvert$ to the proposed neural network in SimGNN. As shown in
Fig. \ref{fig:GraphInteraction}, this module yields the similarity vector to measure the similarity between two graph vectors using a function defined as 
\begin{equation}
f(e_1,e_2)=ReLU(d^TW_1^{[1:S]}d+W_2d+W_3\begin{bmatrix}
     e_1  \\
     e_2 
\end{bmatrix}+b)
\end{equation}
In this equation, $W_1^{[1:S]}\in\mathcal{R}^{S \times F' \times F'}$ is the weight tensor to capture the second-order difference term, $W_2 \in \mathcal{R}^{S\times F'}$ is the weight tensor for first-order difference term, $W_3 \in \mathcal{R}^{S\times 2F'}$ is the weight matrix for the concatenated vector $e_1\|e_2$, $b \in \mathcal{R}^S$ is the learnable bias term and $S$ is the dimension of the similarity vector. 

The similarity vector goes through fully connected layers and a sigmoid function to guarantee that the probability lies in the range of $[0,1]$. By feeding the graph vectors into the graph comparison module, we convert the place recognition to a binary classifier problem and hence we employ binary cross entropy function as the loss for training.  
\begin{equation}
loss=\frac{1}{N_{batch}}\sum^{N_{batch}}_{i=1}y_ilog(\hat{y}_i)+(1-y_i)log(1-\hat{y}_i)
\end{equation}
$y_i\in\{0,1\}$ is the ground truth value, $\hat{y}_i \in [0,1]$ is the prediction value, $N_{batch}$ is the batch size. Lastly, only when the resultant similarity between two graph vectors from the graph comparison module is higher than a threshold, the input scans are passed on to semantic registration module to estimate the 6DoF pose. 


\subsection{Semantic Registration}
\label{RO}

In LiDAR SLAM systems, to reduce the accumulated drift in the front end lidar odometry pipeline, relative pose constraints are added between nodes representing revisted places, and then the whole pose graph with both odometry and loop closure constraints is optimized. Following this, after obtaining potential loop candidates from semantic place recognition module, we estimate the relative pose constraints between potential loop candidates with an enhanced semantic registration algorithm. 

Essentially, to perform robust registration in complex environments, we enhance the front-end pose estimation algorithm in F-LOAM \cite{FLOAM} by combining semantic labels, gaining inspiration from SA-LOAM \cite{SALOAM}. The key enhancements in our semantic registration pipeline are:
\begin{itemize}
    \item Remove outliers and moving/dynamic objects based on the available semantic labels of points.
    \item During data association, associate target points with the same semantic label to construct point to line and point to plane error metrics.
    \item Assign weights based on semantic labels during minimization of the cost function.
\end{itemize}  
First, based on the semantic labels from SemanticKITTI dataset, we remove dynamic objects from the input point cloud such as cars, persons, trucks, bus, other vehicles, outliers and so on. This aids in extracting more robust keypoints, which can alleviate false matching during registration. As in F-LOAM, we then compute the curvature/smoothness and extract a set of edge keypoints ($P_e$) and surface keypoints ($P_s$). Second, the relative pose between two scans is estimated by aligning edge keypoints and surface keypoints using point to line and point to plane error metrics. While calculating the target line and planes corresponding the source edge and planar points, we use nearest 5 points in target scan with the same semantic label. Compared to original F-LOAM, using target points with same semantic labels aids in establishing accurate correspondences. Drawing inspiration from SA-LOAM that different classes influence differently during semantic registration, we assign larger weights to clearly distinguishable classes, such as traffic signs, poles and buildings. For more details, one can look at our implementation in our source code. The cost function is defined as below,
\begin{equation}\label{cost}
r=\sum^{L}_l\left( \sum_i(w^ld^l_{ei})+\sum_j(w^ld^l_{sj})
\right)
\end{equation}
where $d_{ei}^l$ is the distance from the $i$-th edge keypoint to the corresponding edge, $d_{sj}^l$ is the distance from the $j$-th surface keypoint to the corresponding surface, $l \in L$ is the semantic label of the considered keypoint, $w^l$ is the semantic-related weights. We set the weights for traffic signs, poles and buildings label as 1.2 and the rest are set to 0.8. Once the relative pose constraint is calculated, we perform geometric verification based on fitness score, based on which the constraint is added to the pose graph for optimization. 

\section{Experimental Evaluation}\label{sec5}

We design experiments to evaluate the performance of the proposed pipeline and compare with related state-of-the art methods, on open-source datasets. We test its robustness by randomly rotating and occluding input scans while highlighting its low memory footprint and compute requirements. Extensive ablation studies were performed to evaluate multiple ways of encoding geometric information, varying the number of nodes in semantic graph and highlighting the performance improvements contributed by each proposed enhancements. Lastly we evaluate the semantic registration module as a standalone lidar odometry pipeline and later integrate the proposed loop closure module into an open-source SLAM algorithm and present its performance.

\subsection{\textbf{Semantic Place Recognition}}
 We evaluate the performance and robustness of our place recognition module as a classifier, compare it with relevant state of the art methods, and present its memory and compute requirements. 

\subsubsection{Dataset and Implementation Details}
We select three mainstream semantic LiDAR scan datasets, to evaluate our semantic place recognition model's performance. They are SemanticKITTI \cite{SemanticKitti}, KITTI-360 \cite{Kitti360}and KITTI \cite{Kitti} preprocessed by RangeNet++\cite{RangeNet}. SemanticKITTI is the annotated version of KITTI dataset containing 11 publicly open sequences with semantic labels. KITTI-360 is a novel semantic LiDAR dataset which extends SemanticKITTI to much larger areas. Classical KITTI dataset with semantic labels, as produced by RangeNet++ can test our model's robustness when the system emulating real scenarios where the deep learning based segmentation models produce wrong labels. The raw point clouds, available per-point semantic labels and ground-truth poses are used to build datasets to evaluate place recognition. Similar to SGPR, we generate a large set of pairs randomly and if the distance between them is less than 3m, then they are deemed as a true positive pair. If the distance is larger than 20m, they are regarded as a true negative pair. 

Our proposed model is developed using PyTorch using AdamW \cite{AdamW} optimizer with learning rate 0.0001. Our model is trained with batch size of 128 for 50 epochs on one Nvidia Tesla T4 (16 GB). The number of nodes in the semantic graph created from an input scan is by default, set to 50. If the number of segmented instances in a scan are less than 50, then pseudo nodes are added with zero node information. If the semantic instances are more than 50, we randomly sample 50 of them to build the semantic graph to have a consistent batch size for training. In these datasets, most scans contain 30-40 semantic instances and only a few of them go as high as 60-70 nodes per scene graph. Setting the $k$-nearest neighbor parameter $k$ in our GAT to 10 drastically improves the training and inference speed as compared to classical GAT, where all the neighbourhood nodes are used. \textcolor{red}{Different 
$k$ values had negligible impact on the overall performance of the method.}

To make the SemanticKITTI dataset fit in our model, we remap the original 28 classes into 12 appropriate classes (car, other vehicles, other ground, fence, trunk, pole, truck, sidewalk, building, vegetation, terrain and traffic sign), which is the same as SGPR. We select 5 sequences (01, 03, 04, 09, 10) for training and 6 sequences (00, 02, 05, 06, 07, 08) for test. For KITTI-360 dataset, we remap 19 classes into 13 classes(car, static object, ground, parking, rail track, building, wall, fence, guard rail, bridge, pole, vegetation and traffic sign). We train on sequences (00, 02, 03, 04) and test on sequences (05, 06, 07, 09, 10). For KITTI dataset, labels are inferred from the pretrained model of RangeNet++, whose 19 output classes are mapped to 12 classes, same as SemanticKITTI. The train and test sequences are identical to SemanticKITTI. 

\subsubsection{Analysis} We use the maximum value of $F_1$ score as the evaluation metric for our model. It is defined as \[F_1=2\times \frac{P \times R}{P+R}\] where $P$ represents precision and $R$ represents recall. As the place recognition datasets are unbalanced, i.e., the proportion of negative pairs is much larger than positive pairs, the max F1 score is a more comprehensive metric than accuracy(success rate) and average precision. We compare our method with following open-source algorithms SGPR \cite{SGPR}, Scan Context (SC) \cite{ScanContext} and Intensity Scan Context (ISC) \cite{ISC}. While there are other deep-learning-based loop closure algorithms to compare against, most of them need a large amount of work to pre-process the data to get results that can be compared in a fair manner. Hence, we specifically focused on ones with high relevance, leverage semantic graph for place recognition \cite{SGPR} or ones that are widely used such as Scan Context \cite{ScanContext}. Please note that in all the experiments, ``Ours" and ``SGPR" represent our proposed method and SGPR algorithm respectively and the semantic labels are directly taken from the dataset's ground-truth annotations. And when we refer to ``Ours-RN" or ``SGPR-RN", we mimic the real world scenario, where the semantic labels are inferred from a pretrained network, RangeNet++ \cite{RangeNet}, instead of using ground-truth annotations. 
\begin{table*}[]
\centering
\begin{adjustbox}{width=0.95\textwidth}
\begin{tabular}{cccccccc|cccccc}
\hline
\multirow{2}{*}{Method} & \multicolumn{7}{c|}{SemanticKITTI}                                                                                   & \multicolumn{6}{c}{KITTI-360}                                                                       \\ 
                        & 00             & 02             & 05             & 06             & 07             & 08             & Mean           & 05             & 06             & 07             & 09             & 10             & Mean           \\ \hline
SGPR                    & 0.846          & 0.78           & 0.724          & 0.901          & 0.902          & 0.731          & 0.814          & 0.703          & 0.707          & 0.745          & 0.673          & 0.697          & 0.705          \\
SGPR-RN                 & 0.771          & 0.758          & 0.767          & 0.857          & 0.813          & 0.635          & 0.767          & -              & -              & -              & -              & -              & -              \\
SC                      & 0.579          & 0.535          & 0.577          & 0.729          & 0.684          & 0.171          & 0.546          & 0.550          & 0.412          & 0.554          & 0.455          & 0.672          & 0.529          \\
ISC                     & 0.860          & 0.808          & 0.840          & 0.901          & 0.634          & 0.626          & 0.778          & 0.756          & 0.692          & 0.811          & 0.712          & 0.867          & 0.768          \\
Ours-RN                 & 0.923          & 0.839          & \textbf{0.873} & 0.947          & 0.913          & 0.730          & 0.871          & -              & -              & -              & -              & -              & -              \\
Ours                    & \textbf{0.935} & \textbf{0.902} & 0.858 & \textbf{0.979} & \textbf{0.926} & \textbf{0.923} & \textbf{0.921} & \textbf{0.816} & \textbf{0.824} & \textbf{0.890} & \textbf{0.762} & \textbf{0.916} & \textbf{0.842} \\ \hline
\end{tabular}
\end{adjustbox}
\caption{Evaluation of various methods on SemanticKITTI and KITTI-360 dataset using max F1 score metric. Ours-RN and SGPR-RN represent the performance of our proposed system and SGPR with semantic labels inferred from RangeNet++ on the KITTI dataset. }
\label{tab:ThreeResult}
\end{table*}

\subsubsection{Evaluation using Max F1 Score}
\label{sec:eval}
In Table. \ref{tab:ThreeResult}, we compare the max F1 score of our proposed method (Ours) and Ours-RN (semantic labels inferred from RangeNet++) with SC \cite{ScanContext}, ISC \cite{ISC}, SGPR \cite{SGPR} and \mbox{SGPR-RN} (semantic labels inferred from RangeNet++). Table \ref{tab:ThreeResult} shows that our proposed method (Ours) achieves the highest performance of 0.921 and 0.842, when compared to other approaches on all sequences on both datasets. This translates to a 13\% and 19\% improvement over baseline SGPR algorithm on SemanticKITTI and KITTI-360 datasets respectively. 
We can also notice that the max F1 score in general is lesser on KITTI-360 dataset in comparison to SemanticKITTI dataset. The reason for this is that the annotation of KITTI-360 is provided in submaps of combined scans and we had to use a clustering method to recover labels for each scan causing some annotation errors/noise. The performance of Ours-RN and SGPR-RN are missing on KITTI-360 sequences as they contain about 25000 frames per sequence, which requires weeks of pre-processing to get RangeNet++ and our evaluation results. Hence, we used the available ground-truth annotations and evaluated them directly on KITTI-360 dataset.

\subsubsection{Evaluation with real-world semantic label inference}
The results of Ours-RN and SGPR-RN in Table. \ref{tab:ThreeResult} refer to the performance on KITTI dataset with semantic labels inferred from pretrained RangeNet++ model. It can be clearly seen that inference with pretrained model (Ours-RN and SGPR-RN) offers slightly lower performance than their counterparts with ground-truth semantic annotations (Ours and SGPR). This is expected behaviour as inference with a pre-trained deep learning model, in this case, RangeNet++ can never reach the accuracy of ground-truth annotations. However, the drop in loop closure performance is not significant (about 5\% max F1 score), when compared to the drop in RangeNet++ performance to groundtruth annotations (about 48\% mean IoU score). Ours-RN still maintains high performance even with pre-trained model to infer semantic labels, mainly because of its architectural advantage of using a semantic graph representation, which is tolerant to noisy labels in few graph nodes, while maintaining a distinctive high-level semantic graph representation of every scene.

\subsubsection{Evaluation using Precision-Recall curves}
\begin{figure*}[ht]
    \centering
    \begin{tabular}{ccc}
        \includegraphics[width=0.33\linewidth]{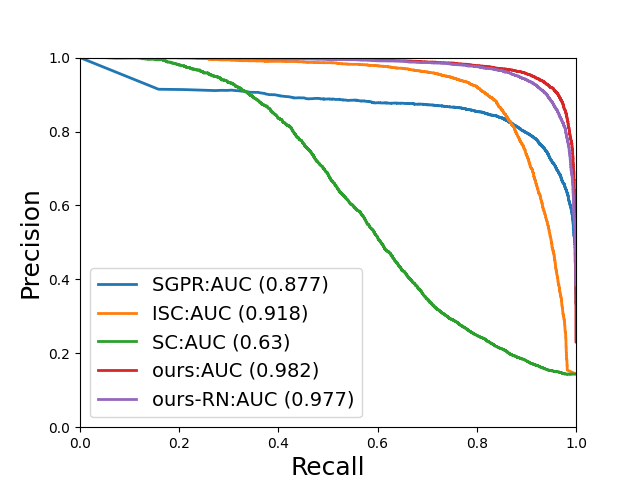} &
        \includegraphics[width=0.33\linewidth]{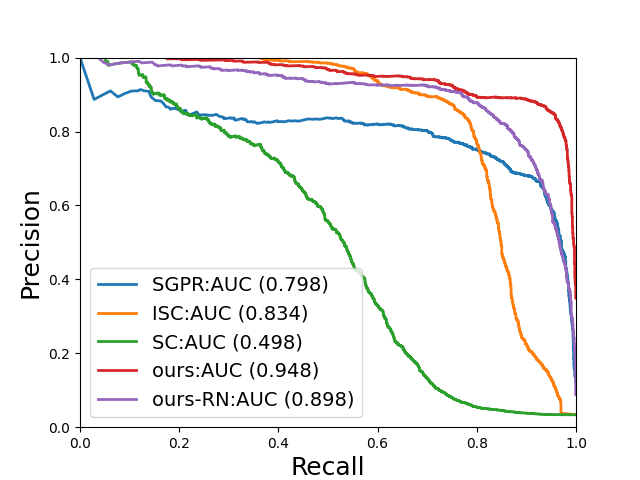} &
        \includegraphics[width=0.33\linewidth]{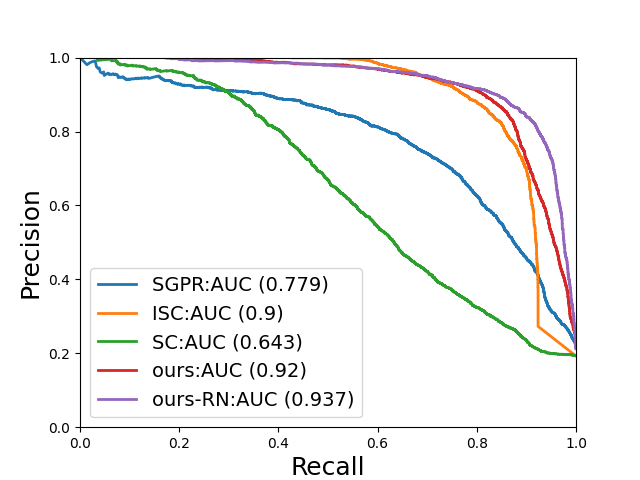} \\
        \small{(a) Sequence-00} &
        \small{(b) Sequence-02} &
        \small{(c) Sequence-05} \\
        \includegraphics[width=0.33\linewidth]{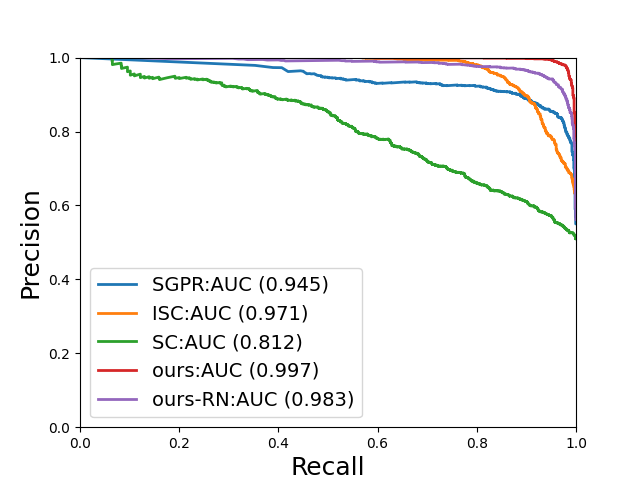} &
        \includegraphics[width=0.33\linewidth]{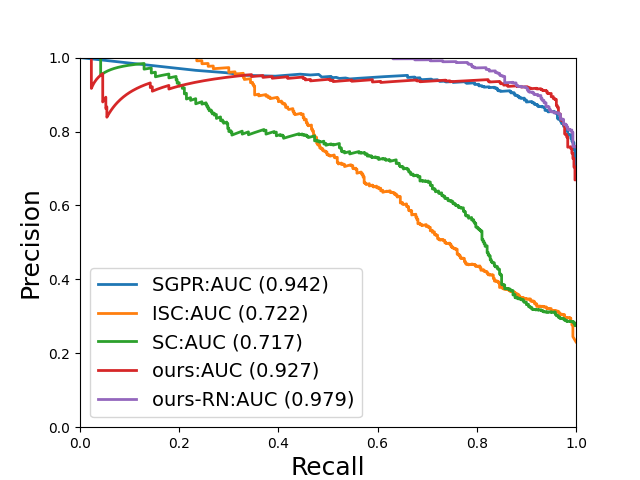} &
        \includegraphics[width=0.33\linewidth]{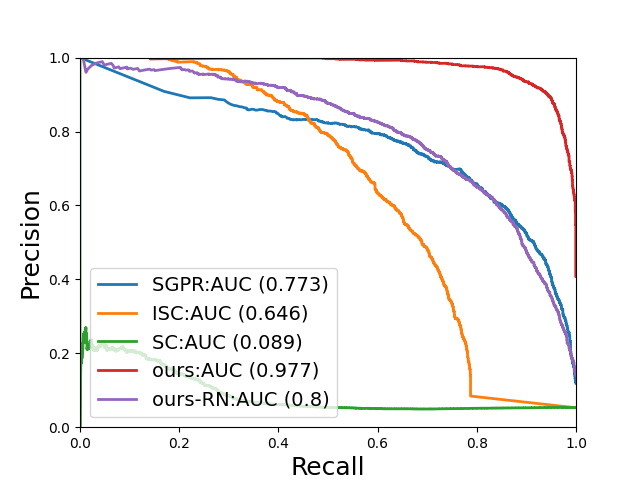}\\
        \small{(d) Sequence-06} &
        \small{(e) Sequence-07} &
        \small{(f) Sequence-08} \\
    \end{tabular}
    \caption{Precision-Recall curves of max F1 score on SemanticKITTI dataset. In the legend, AUC denotes the area under curve. Here we compare, Ours and Ours-RN with SGPR \cite{SGPR}, SGPR-RN, ScanContext (SC) \cite{ScanContext} and ISC \cite{ISC}. It can be seen that Ours and Ours-RN outperform other methods on all sequences, and especially on Sequence 08, where there are many reverse loop closures.}
    \label{fig:PR}
\end{figure*}

 In Fig. \ref{fig:PR}, we show the precision recall curves of max F1 score on various sequences of SemanticKITTI dataset. It can be seen that on most sequences, the area under curve (AUC) or average precision of our model (Ours and Ours-RN) is much larger than the baseline SGPR or widely used ScanContext. Especially in sequence 08, there are some reverse loop closures, wherein the robot visits the same place but in reverse direction. The reverse loop closure detection is a challenging problem for most existing loop closure detection solutions and most traditional methods fail to recognize such loops. However, our model combining contextual information via GAT can discern the place by spatial topology and semantic understanding. Hence, it can be seen that in Fig.\ref{fig:PR} (f), which corresponds to sequence 08, Ours and Ours-RN offer significantly better performance than other methods.

\begin{table*}[]
\centering
\begin{adjustbox}{width=0.95\textwidth}
\begin{tabular}{ccccccccc|cccccccc}
\hline
\multirow{2}{*}{Method} & \multicolumn{8}{c|}{Rotation}                                                                                                         & \multicolumn{8}{c}{Occlusion}                                                                                                          \\ 
                        & 00             & 02             & 05             & 06             & 07            & 08             & Mean           & CMP             & 00             & 02             & 05             & 06             & 07             & 08             & Mean           & CMP             \\ \hline
SGPR                    & 0.741          & 0.701          & 0.708          & 0.905          & 0.734         & 0.675          & 0.744          & -0.070          & 0.815          & 0.672          & 0.721          & 0.927          & 0.894          & 0.695          & 0.787          & \textbf{-0.027}          \\
SC                      & 0.269          & 0.112          & 0.323          & 0.668          & 0.386         & 0.192          & 0.325          & -0.221          & 0.524          & 0.447          & 0.530           & 0.654          & 0.158          & 0.386          & 0.450          & -0.096          \\
ISC                     & 0.857          & 0.804          & 0.836          & 0.899          & 0.626         & 0.627          & 0.775          & -0.003          & 0.833          & 0.793          & 0.810           & 0.878          & 0.578          & 0.600          & 0.749          & -0.029 \\
Ours                    & \textbf{0.927} & \textbf{0.898} & \textbf{0.877} & \textbf{0.982} & \textbf{0.910} & \textbf{0.918} & \textbf{0.919} & \textbf{-0.002} & \textbf{0.923} & \textbf{0.831} & \textbf{0.836} & \textbf{0.972} & \textbf{0.846} & \textbf{0.802} & \textbf{0.868} & -0.052          \\ \hline
\end{tabular}
\end{adjustbox}
\caption{Robustness evaluation using random rotation and occlusion on SemanticKITTI. The “CMP” column shows the difference between the mean values of corresponding algorithms from this Table. \ref{tab:RotAndOcc} and Table. \ref{tab:ThreeResult}. Our method shows excellent robustness and has the lowest drop in performance when compared to the performance on original SemanticKITTI dataset.}
\label{tab:RotAndOcc}
\end{table*}

\subsubsection{Robustness to random rotation and occlusion}





In order to test robustness in real-world scenarios, we randomly rotate and occlude the the scans in SemanticKITTI dataset. Specifically, we apply a random yaw angle between $-30^{\circ}$ to $30^{\circ}$ and mask some points in a randomly generated horizontal field of view of $30^{\circ}$. We perform the same experiments as in Section. \ref{sec:eval} to estimate max F1 score after this random rotation and occlusion, and the resulting performance is shown in Table. \ref{tab:RotAndOcc}. Compared to other methods, Table. \ref{tab:RotAndOcc} shows that our proposed algorithm offers the highest max F1 score on all sequences. 

In Table. \ref{tab:RotAndOcc}, the ``CMP" column shows the difference between the mean values of corresponding algorithms from Table. \ref{tab:RotAndOcc}. and Table. \ref{tab:ThreeResult}. Basically, it shows that among all methods, our proposed algorithm has the lowest drop in mean performance when compared to its original performance in Table. \ref{tab:ThreeResult}. The reason being, that semantic graph representation essentially encodes the semantic structure of the scene in a sparse manner, making it robust to occlusion of objects and small variations arising from wrong semantic label inference. This particularly makes our proposal an ideal choice for loop closure detection in real-world scenarios. 

\subsubsection{Model size \& Computational time for inference}
\begin{table}[hbtp]
\centering
\begin{adjustbox}{width=0.6\textwidth}
\begin{tabular}{cc}
\hline
Method       & Model Size \\ \hline
MinkLoc \cite{Mink}      & 4MB        \\
LCD-Net \cite{LCDNet}     & 142MB      \\
PointNetVlad \cite{PointNetVLAD}& 237MB      \\
LPD-Net \cite{LPDNet}& 197MB      \\
Ours         & 426KB      \\ \hline
\end{tabular}
\end{adjustbox}
\caption{Model size comparison.} 
\label{tab:Size}
\end{table}
One of the impressive features of our proposed model is that its extremely lightweight in terms of memory, which inturn makes it extremely fast. As shown in Table. \ref{tab:Size} our model requires only 426 KB, which is extremely less as compared to other learning-based methods. This minimal model size makes our model trainable and deployable even on a consumer laptop. Owing to its small model size, our proposed model runs extremely fast, at about 73 Hz, on an Nvidia Tesla T4 (15 GB) GPU for inference with pre-processed semantic graphs as input. Our proposed model runs in sequential mode to instance segmentation algorithm, and most state-of-the-art instance segmentation algorithms run at around 1-5 Hz depending on the hardware and their level of optimization. This shows that our model has minimal computational requirements and does not slow down the system by becoming a bottleneck.

\subsection{\textbf{Ablation Study}}
\label{sec:ablation}

We perform three ablation studies, first, evaluate multiple ways of encoding geometric information, second, varying the number of nodes in semantic graph and lastly, highlighting the performance improvements that each proposed enhancements bring to the whole pipeline.
\begin{figure}
  \begin{center}
  \includegraphics[width=0.7\linewidth]{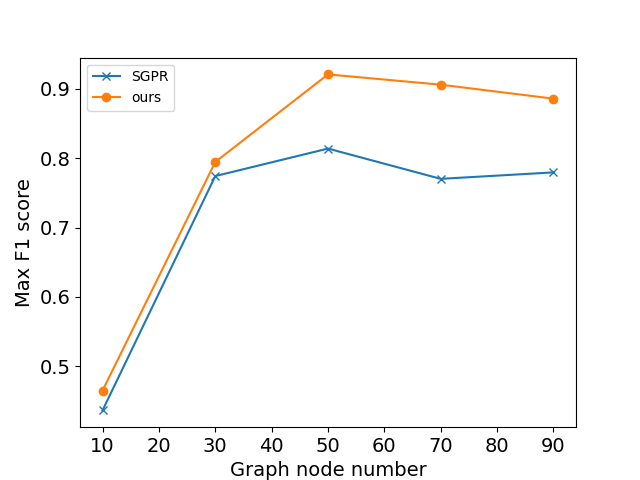} 
  \end{center}

  \caption{Performance change with varying number of nodes in semantic graph. }
  \label{fig:NodeCompare}
\end{figure}

\subsubsection{Evaluation with multiple geometric features}
\label{sec:geo_ablation}
There are multiple ways of encoding geometric information in the semantic graph, and we tested three options to find out the best performing one with our pipline. First, encoding the 3D neighbourhood using one of best classical 3D feature descriptor, FPFH \cite{fpfh}. Second, we integrated the features extracted using a pretrained PointNet \cite{PointNet} model and lastly we used the bounding box information that's readily available from instance segmentation. While there are other 3D feature descriptors such as SHOT \cite{shot} and 3DHoPD \cite{3dhopd}, we chose FPFH owing to its low dimensionality and ease of use. For example, SHOT is a 352 dimensional descriptor while FPFH is just 33 dimensional and 3DHoPD's design requires a two-stage feature extraction and matching, making FPFH an easier choice to integrate. 

To extract FPFH features, we operate on each instance point cloud, by considering it's centroid as the keypoint, feature radius as the maximum distance from centroid to the farthest point, while extracing normals from 30 nearest neighbours. For PointNet features, we used the pretrained model for classification task on ModelNet40 \cite{modelnet40} and the dimension of output global feature is 1024. Lastly for bounding box, we just use the top left and bottom right 3D points forming a six dimensional vector. The resulting max F1 score on SemanticKITTI dataset, with these three different geometric features is shown in Table \ref{tab:geoCompare}. It clearly highlights that, though intuitively, classical and deep learning descriptors encode more information about neighbourhood variance and point statistics, it does not essentially improve the whole system performance, when added to the semantic graph along with spatial and semantic labels. Instead, the lightweight bounding boxes that essentially just capture the boundaries of the objects and more importantly, their scale information representing how big the objects are, is able to aid loop closure detection in a much better way, offering higher performance.
\begin{table*}[]
\centering

\begin{tabular}{ccc}
\hline
Geometry Feature & Dimension & Mean  \\ \hline
Bounding box     & 6         & \textbf{0.921} \\
FPFH             & 33        & 0.876 \\
PointNet         & 1024      & 0.883 \\ \hline
\end{tabular}
\caption{Maximum F1 score comparison of different geometric features on SemanticKITTI dataset.}
\label{tab:geoCompare}
\end{table*}

\subsubsection{Impact of number of nodes in semantic graph}

In all our experiments, we set the maximum number of nodes in the semantic graph to \textit{50}. If there are less nodes, we set them to \textit{zero} and if there are more nodes, we randomly sample \textit{50} from them. In most cases, on existing datasets, the number of nodes were less than 50, with a few samples ranging around 60 or more. In this experiment, we test the influence of number of graph nodes on the model's performance. The Fig. \ref{fig:NodeCompare} compares the performance of SGPR with our model, as the graph node number varies from 10 to 90. It can be seen that our model always performs better than SGPR with same number of graph nodes. When there are less number of nodes \textit{(10-30)}, the semantic graph discards most of the semantic information and encodes a partial set resulting in lower performance. It can be seen that choosing 50 or more number of graph nodes offers almost similar performance, with negligible change in performance.

\subsubsection{Performance improvement from proposed enhancements}
Here, we show how each of our proposed enhancements bring considerable improvement to the baseline SGPR algorithm using the SemanticKITTI dataset. We dissect our complete system, starting from SGPR as baseline and add one module after another, and show the resulting improvement in max F1 score in Table \ref{tab:ablation}. The baseline SGPR is first enhanced with DIFF module, which stands for the relative difference term which we added in the graph comparison module, resulting in a jump from 0.814 to 0.876 (~8\%). We then replace EdgeConv in SGPR with graph attention networks (GAT) and self-attention that immediately follows it in Fig. \ref{fig:NodeEmbedding}, to encode complex graph relations into feature vectors, demonstrating further improvement in the performance. We then add geometric features (GEO), i.e., bounding box information which supplements the graph with scale information and boundaries of detected objects, resulting in further performance enhancement. Lastly, we show that using self-attention (ATT) in graph embedding to create a representative global context vector, as discussed before in graph embedding part, again improves the performance by another 2\%, as shown in Table \ref{tab:ablation}. 
\textcolor{red}{We also analyzed the impact of each individual module in our pipeline. The results indicate that the GAT, GEO, and ATT modules not only enhance performance when combined but also contribute to accuracy improvements independently. Furthermore, the findings confirm that the DIFF serves a dominant role in the overall performance enhancement of the proposed method.}
\begin{table}[]
\centering
\begin{adjustbox}{width=0.6\textwidth}
\begin{tabular}{ccccc|c}
\hline
SGPR       & DIFF       & GAT        & GEO        & ATT        & Mean           \\ \hline
\checkmark &            &            &            &            & 0.814          \\
\checkmark & \checkmark &            &            &            & 0.876          \\
\checkmark & \checkmark & \checkmark &            &            & 0.882          \\
\checkmark & \checkmark & \checkmark & \checkmark &            & 0.908          \\
\checkmark & \checkmark & \checkmark & \checkmark & \checkmark & \textbf{0.921} \\ \hline
\checkmark &   & \checkmark &            &            & \textcolor{red}{0.838}          \\
\checkmark &  &  & \checkmark &            & \textcolor{red}{0.829}          \\
\checkmark &  &  &  & \checkmark & \textcolor{red}{0.834} \\ \hline
\end{tabular}
\end{adjustbox}
\caption{Ablation study on SemanticKITTI dataset showing how each proposed enhancement to baseline SGPR contributes to overall performance improvement of the proposed place recognition module. SGPR denotes the baseline model, DIFF denotes the relative difference term added in the graph comparison module of Fig. \ref{fig:GraphInteraction}, GAT stands for the graph attention networks added in semantic graph encoder along with the immediate self-attention module to generate node embedding $f$ (refer Fig. \ref{fig:NodeEmbedding}), GEO represents the geometric information branch that encodes the bounding box information, and ATT is the self-attention layer from the graph embedding part that creates the global context vector $c$ (refer Fig. \ref{fig:NodeEmbedding}).}
\label{tab:ablation}
\end{table}

\subsection{\textbf{Semantic Registration}}
\label{odom}

\subsubsection{LiDAR odometry estimation with Semantic registration} 

Our semantic registration is an enhanced front-end pose estimation algorithm from F-LOAM \cite{FLOAM}, with semantic label assisted dynamic point removal, correspondence association and weighting. As our semantic registration algorithm can also function as a front-end LiDAR odometry algorithm, we evaluate its pose estimation accuracy on SemanticKITTI dataset. We compare vanilla F-LOAM and our enhanced version with semantic labels, based on their odometry accuracy on multiple sequences, as shown in Table. \ref{tab:odometry}. The code is implemented using ROS \cite{ros} in C++ with an Intel Core i7-7700HQ CPU (2.80GHz × 8). We employ the absolute trajectory error (ATE) \cite{ATE} to evaluate the accuracy of estimated trajectory. For this, the error matrix $E$ at time $i$ is defined as as 
\[
E_i=Q_i^{-1}SP_i
\]
where $Q_i \in \mathit{SE}(3)$ is the ground-truth pose, $P_i \in \mathit{SE}(3)$ is the estimated pose, $S$ is the rigid body transformation matrix. Then ATE is calculated as the root mean square error from error matrices.
\[
\text{ATE}=\sqrt{\left( 
\frac{1}{n} \sum_{i=1}^n \| E_i^T\|^2
\right)}
\]
ATE is reflective of the average deviation between the estimated pose and ground-truth pose per frame. 

\begin{table}[]
\centering
\begin{adjustbox}{width=0.75\textwidth}
\begin{tabular}{cccccc}
\hline
Method & 00             & 02             & 05             & 08             & 09             \\ \hline
F-LOAM    & 4.998          & 8.388          & \textbf{2.854} & 4.112          & 1.806          \\
Ours-odometry   & \textbf{4.590} & \textbf{8.378} & 3.229          & \textbf{3.734} & \textbf{1.175} \\ \hline
\end{tabular}
\end{adjustbox}
\caption{Comparison between F-LOAM and semantic registration based odometry using ATE metric on SemantiKITTI dataset. This shows that semantic registration is accurate and can work on long sequences.}
\label{tab:odometry}
\end{table}
We select 5 sequences (00, 02, 05, 08, 09) to compare these two methods. From Table \ref{tab:odometry}, we can see that our semantic-assisted registration generates more accurate trajectories than the original F-LOAM on four sequences except on sequence 05. The possible reason for this can be attributed to few instances where the pose estimation was erroneous, which then got propagated to later frames, as its an open-ended odometry system with no loop closure. We want to highlight that the proposed semantic registration is often more accurate (our first requirement) and robust enough to work on complete sequences without loop closure. Please do not mistake this for complete system accuracy with loop closure, which is covered in the next experiment. Even after integrating semantic information, the semantic registration algorithm runs at 9.3 Hz, while vanilla F-LOAM runs at 10.2 Hz on above system configuration, showing negligible difference.

\begin{figure}[ht]
  \begin{center}
  \includegraphics[width=0.7\linewidth]{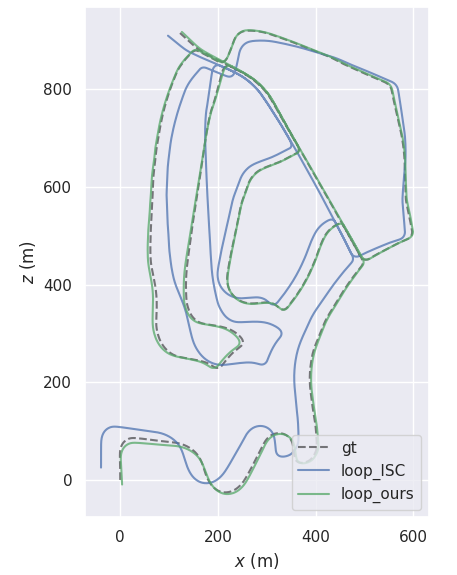} 
  \end{center}

  \caption{Trajectories of complete SLAM system based on our loop closure module, ISC-LOAM and the ground truth of sequence 02 from SemanticKITTI.}
  \label{fig:loopcompare}
\end{figure}

\begin{table}[]
\centering
\begin{adjustbox}{width=0.64\textwidth}
\begin{tabular}{ccccc}
\hline
Method   & 00             & 02             & 05             & 08             \\ \hline
ISC-LOAM & 1.304          & 39.114         & 0.730 & 3.860          \\
Ours     & \textbf{1.252} & \textbf{3.213} & \textbf{0.689} & \textbf{3.699} \\ \hline
\end{tabular}
\end{adjustbox}
\caption{Comparison of our proposed loop closure algorithm and intensity scan context based ISC-LOAM with same front-end algorithm, F-LOAM, on SemanticKITTI dataset. 
}
\label{tab:loop}
\end{table}
\subsection{\textbf{Semantic Loop Closure}}

We finally compare our whole semantic loop closure system consisting of the proposed semantic place recognition module and semantic registration to estimate the 6DoF pose, by plugging it into ISC-LOAM's framework. We compare our proposed back-end module with ISC-LOAM's back-end that uses intensity scan context, while keeping the front-end odometry algorithms unchanged. 

We conducted the experiments on SemanticKITTI's 4 sequences (00, 02, 05, 08) as they contain enough loops, following the settings described in Section. \ref{odom} with ATE as the evaluation metric. The results are shown in Table. \ref{tab:loop} conclude that our semantic loop closure algorithm finds enough loops and semantic registration estimates accurate relative pose constraints, resulting in superior performance over the traditional LiDAR SLAM, ISC-LOAM. Especially on sequence 02 (Fig. \ref{fig:loopcompare}), ISC-LOAM fails to close the loop, may be because of its wrong loop closure detection or inaccurate relative constraints, while our proposed method runs successfully. Thus our proposed loop closure detection algorithm based on semantic graph encoder and graph comparison module is a robust and reliable alternative with minimal memory and computational requirements.

\section{Conclusion \& Future Work}\label{sec6}
We introduced a LiDAR-based loop closure detection algorithm that uses semantic graphs with graph attention networks to identify possible revisited places. Our algorithm had two modules, namely the semantic graph encoder module and the graph comparison module. The semantic graph encoder encodes the spatial, semantic and geometric information from the scene graph of point clouds using graph attention neural networks and self-attention mechanism to create distinctive graph vectors. These graph vectors of candidate loop closure scans are then classified as a successful match or not by the graph comparison module, mainly leveraging on the difference of these graph vectors in an end-to-end trainable network. Lastly, we implemented a semantic registration algorithm to estimate the 6 DoF pose and integrated it into existing LiDAR SLAM algorithm. Our experiments show that the proposed approach offers a significant boost in performance and opens up a direction of employing graph attention networks for improved accuracy in place recognition. Lastly, to foster further research in this direction, we open-source our complete algorithm. 

\textcolor{red}{Our model is limited by predefined semantic labels for training. Exploring the direction of leveraging foundational models such as SAM \cite{SAM}, CLIP \cite{CLIP} and LLMs \cite{3dllm} to automatically segment and reason about the point cloud information is helpful for in-the-wild depolyment. In the future, we also plan to employ RGB information, leverage classical (learning-free) loop closure detection algorithms and develop an even more efficient algorithm to strike a balance between the accuracy and run time of our module.} 

\backmatter







\section*{Declarations}
\subsection{Acknowledgements}
Not Applicable

\subsection{Funding}
Not Applicable

\subsection{Conflict of interest/Competing interests}
The authors declare that they have no conflicts of interest or competing interests.

\subsection{Ethics approval}
Not applicable

\subsection{Consent to participate}
Informed consent was obtained from all individual participants included in the study.

\subsection{Consent for publication}
Consent for publication was obtained from all participants whose data is included in this manuscript.
\subsection{Data and code availability} 
The raw data that support the findings of this study are openly available at website of SemanticKITTI and KITTI-360. The custom code used for data analysis is available on GitHub at https://github.com/crepuscularlight/SemanticLoopClosure.




\bibliographystyle{sn-basic}
\bibliography{sn-bibliography}

\end{document}